\documentclass{article}

\PassOptionsToPackage{square,numbers}{natbib}

\usepackage[preprint]{neurips_2023}

\usepackage[utf8]{inputenc} 
\usepackage[T1]{fontenc}    
\usepackage[hidelinks]{hyperref}       
\usepackage{url}            
\usepackage{booktabs}       
\usepackage{amsfonts}       
\usepackage{nicefrac}       
\usepackage{microtype}      
\usepackage{xcolor}         

\usepackage{graphicx}
\usepackage{amsmath}

\usepackage[inline]{enumitem}
\usepackage{outlines}
\newlist{todo}{itemize}{4}  
\setlist[todo,1]{label=$\square$}
\setlist[todo,2]{label=\normalfont\bfseries \textendash}
\setlist[todo,3]{label=\textasteriskcentered}
\setlist[todo,4]{label=\textperiodcentered}
\usepackage{pifont}
\newtheorem{remark}{Remark}[section]

\newcommand{\nans}{\texttt{NaN}s}
\newcommand{\sign}{\textit{S}}
\newcommand{\signgx}{\sign(g(x))}

\newcommand{\titletext}{Software-based Automatic Differentiation is Flawed}
\title{\titletext}

\providecommand{\stanford}{Stanford University}

\author{%
  Daniel Johnson \\
  \stanford \\
  \texttt{dansj@stanford.edu} \\
  \And 
  Trevor Maxfield \\
  \stanford \\
  \texttt{maxfit@standford.edu} \\
  \And 
  Yongxu Jin \\
  \stanford \\
  \texttt{yxjin@stanford.edu} \\
  \And
  Ronald Fedkiw \\
  \stanford \\
  \texttt{rfedkiw@stanford.edu} \\
}

\author{%
Daniel Johnson \quad Trevor Maxfield \quad Yongxu Jin \quad Ronald Fedkiw \\
Stanford University \\
\texttt{\{dansj,maxfit,yxjin,rfedkiw\}@stanford.edu}
}

\begin{document}

\maketitle

\begin{abstract}
Various software efforts embrace the idea that object oriented programming enables a convenient implementation of the chain rule, facilitating so-called automatic differentiation via backpropagation.  Such frameworks have no mechanism for simplifying the expressions (obtained via the chain rule) before evaluating them.  As we illustrate below, the resulting errors tend to be unbounded.
\end{abstract}

\section{Introduction}
Basic calculus texts introduce students to many of the rules of differentiation including the product rule, quotient rule, chain rule, etc.
Various large-scale software packages (e.g.,~PyTorch \cite{paszke2019pytorch}, TensorFlow \cite{abadi2016tensorflow}, Torch \cite{torch}, Caffe \cite{jia2014caffe}, Theano \cite{al2016theano}, Jax \cite{jax2018github}) utilize the various rules of differentiation, leveraging the idea of object oriented programming in order to elegantly implement the chain rule.
Each implemented function is paired with code that computes its derivative, and branches are created to connect functions that call other functions.
Using the values obtained in a so-called forward pass through the code, the computational graph is evaluated by following the branches in order to obtain values for various derivatives.
As is taught (as early as the discussion on limits) in various calculus texts, one ought to simplify expressions before evaluating them. 
This is even more important when considering how the rules of differentiation (especially the chain rule) increase the number of expressions.
There seems to be no good remedy for expression simplification in the popular software packages.

\section{A Trivial Example}
\begin{figure}[p]
	\centering
 \includegraphics[width=\linewidth]{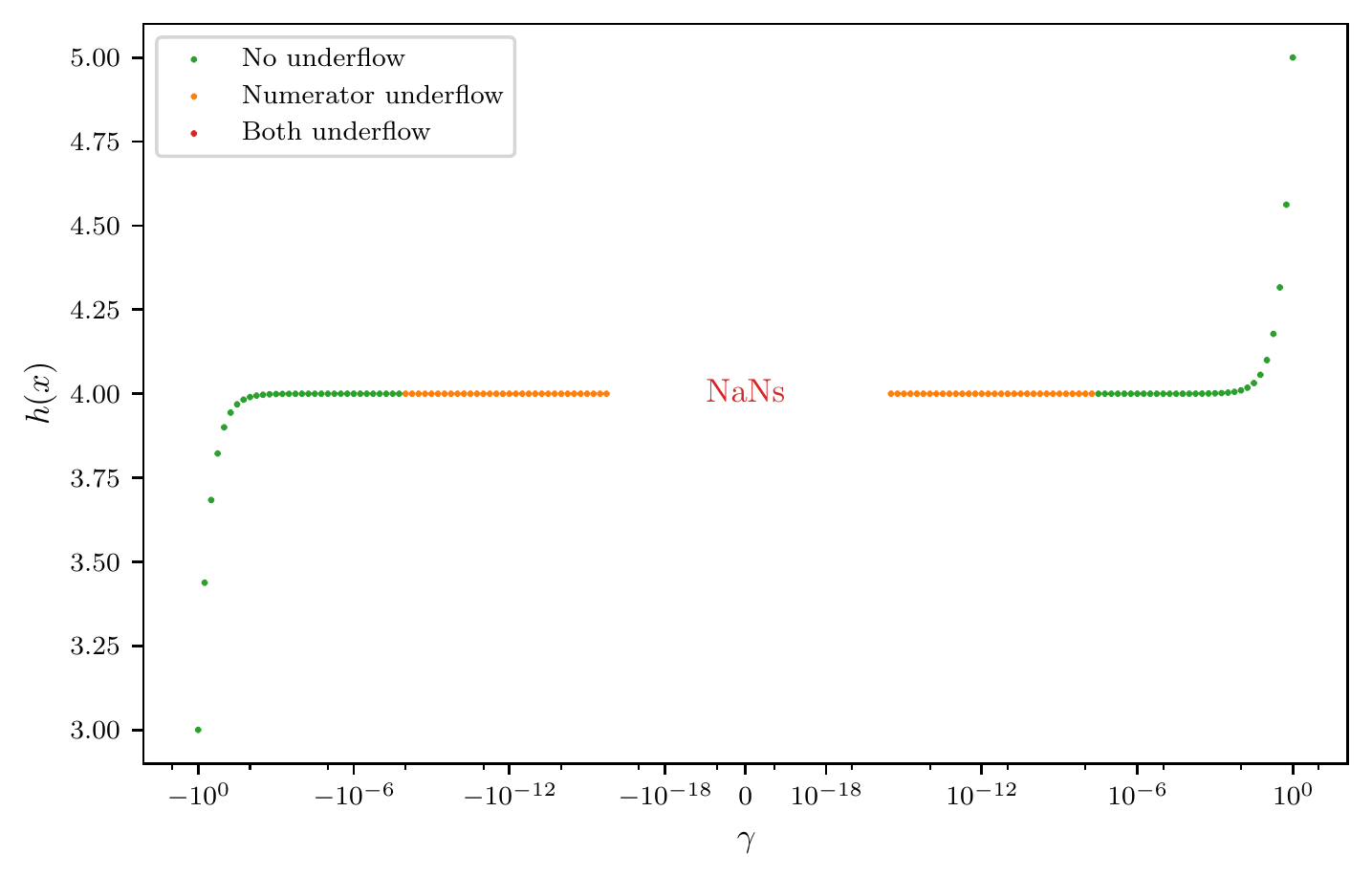}
	\caption{To understand the behavior of $h(x)$ near $x \approx 2$, we choose $\gamma$ with $\gamma \approx 0$ and evaluate $x=2+\gamma$ before subsequently evaluating $h(x)$.  The plot shows values obtained for $h(x)$ with $\gamma$ plotted on a log scale. The green points are consistent with the straight line 
 $h(2+\gamma)=\gamma+4$ as expected. The yellow points indicate values of $\gamma$ where the numerator underflows, i.e. where $x^2 = 4 + 4\gamma + \gamma^2$ evaluates to $4 + 4 \gamma$ because $\gamma^2 \ll 4$ using double precision.  With a numerator of $4 + 4\gamma - 4 = 4\gamma$ and a denominator of $2 + \gamma - 2 = \gamma$, the yellow points are consistent with $h(2+\gamma) = 4$.  The empty space near $\gamma=0$ is where the denominator also underflows, i.e. when $x=2+\gamma$ evaluates to 2 because $\gamma \ll 2$ using double precision.  These values of $\gamma$ produce \nans.  See also Figure~\ref{fig:h-4}.}
	\label{fig:h}
\end{figure}

\begin{figure}[p]
	\centering
 \includegraphics[width=\linewidth]{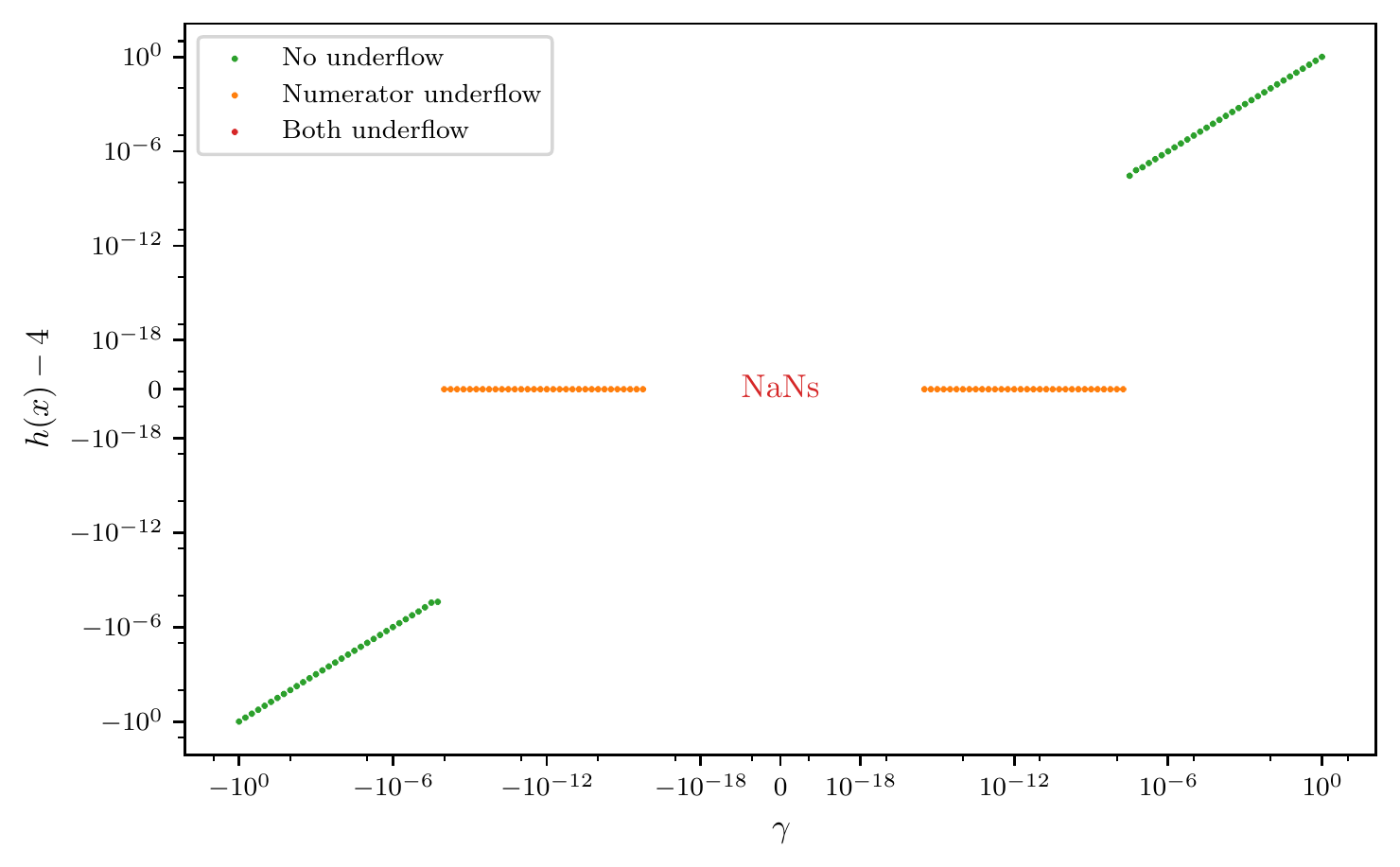}
	\caption{Same data as in Figure~\ref{fig:h}, except that the leading digit of 4 is removed from all of the data.  Plotting the dependent axis on a log scale elucidates the linearity in the green region and the incorrect flatness in the yellow region.  In addition, note the small errors in the linearity of the green region near the boundary with the yellow region. These errors are caused by $x^2 = 4 + 4\gamma + \gamma^2$ evaluating to $4 + 4\gamma + \alpha \gamma^2$ as information in $\gamma^2 \ll 4$ is partially lost as it begins to underflow. Note that $0 < \alpha < 1$ is what one would expect when some of $\gamma^2$ is truncated; however, $\alpha$ can be larger than 1 when rounding up before truncating.  Generally, one would expect $1 < \alpha < 2$ when rounding up, since the largest increase occurs when $2^{-p}$ is rounded up to $2^{-p+1}$.
    }
	\label{fig:h-4}
\end{figure}

Suppose one simple function is $f(x)= x^2-4$ and another is $g(x) = x-2$; then $h(x) = \frac{f(x)}{g(x)}$ has $h(2) = \texttt{NaN}$ on an AMD Ryzen 5 5600X 6-Core Processor, Apple M1 Pro, etc. (any system correctly implementing IEEE 754 \cite{IEEE_754_2019}).  Figures~\ref{fig:h} and \ref{fig:h-4} show $h(x)$ near $x=2$.

\begin{remark}
Replacing the numerator of $h$ with $x^3-8$ causes it to underflow in a wider interval.  The higher the powers of $x$, the larger the interval of underflow.  The width of the domain where the denominator leads to \nans\ can likewise be increased.  
\end{remark}

\begin{figure}[h]
	\centering
\includegraphics[width=0.9\linewidth]{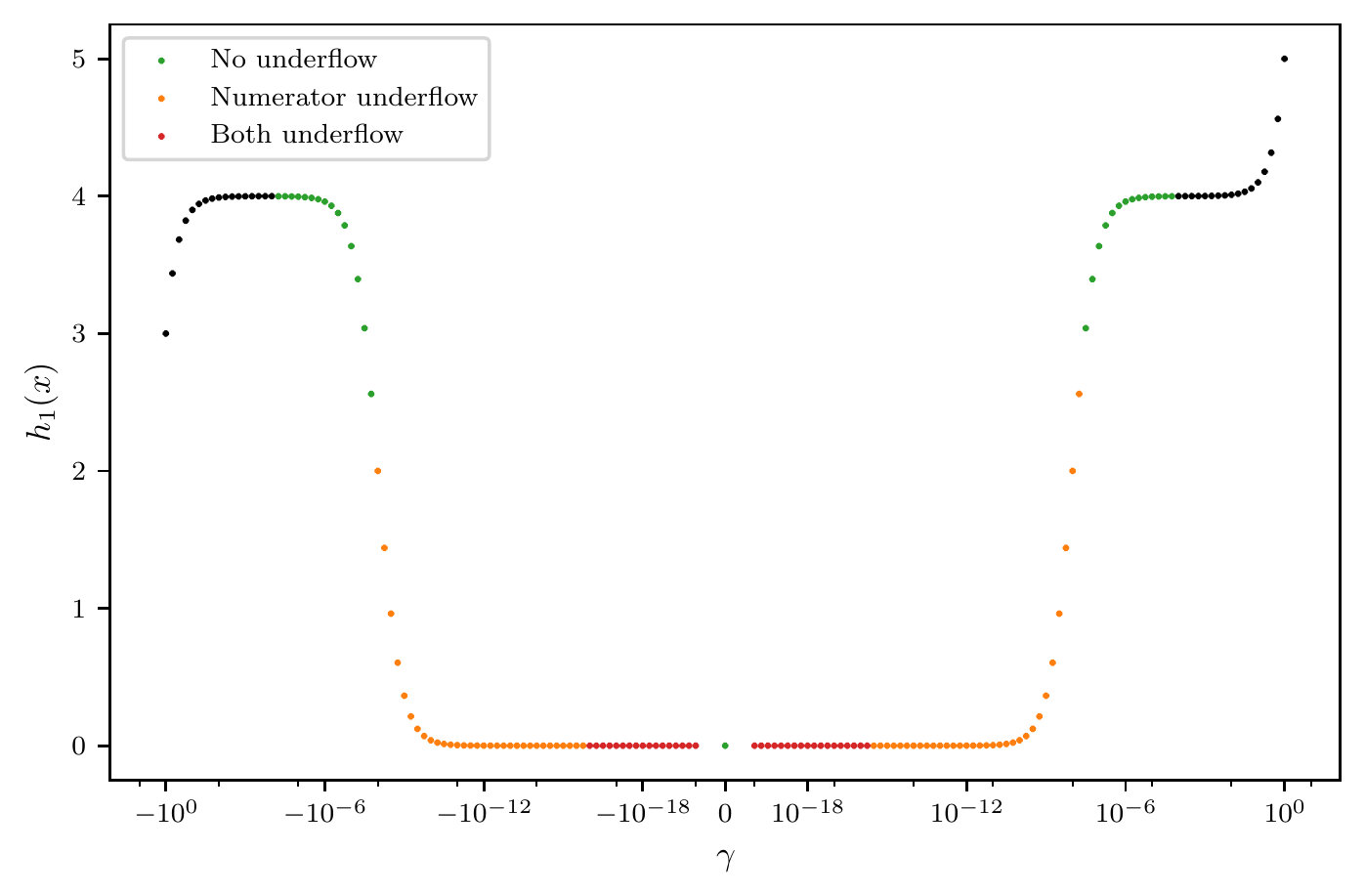} \\
	\caption{
Plot of $h_1(2+\gamma)$ with $\epsilon = 10^{-8}$ and $\delta = 10^{-4}$, where $\delta$ is chosen large enough to put the discontinuity in the green region, well-separated from the issues with underflow we wish to address.
The black points indicate data where $h_1$ is identical to $h$. 
Although the perturbed denominator removes the \nans~in the red region, $h_1$ is identically zero there and thus inconsistent with $x+2$;
moreover, the yellow and green points are also incorrectly perturbed away from $x+2$ towards zero.
These $O(1)$ errors are difficult to avoid without refactoring the code.
Note that the $\gamma=0$ data point is colored green because $2+\gamma$ does not technically underflow (even though $2+\gamma$ evaluates to $2$, and thus $\gamma=0$ behaves similarly to the red points); in addition, the gap around $\gamma = 0$ is due to the minimum $|\gamma|$ being $10^{-20}$ (besides $\gamma =0$).
See also Figure \ref{fig:h_1_84_y-4}.}
	\label{fig:h_1_84_y}
\end{figure}

\begin{figure}[h]
	\centering
\includegraphics[width=0.9\linewidth]{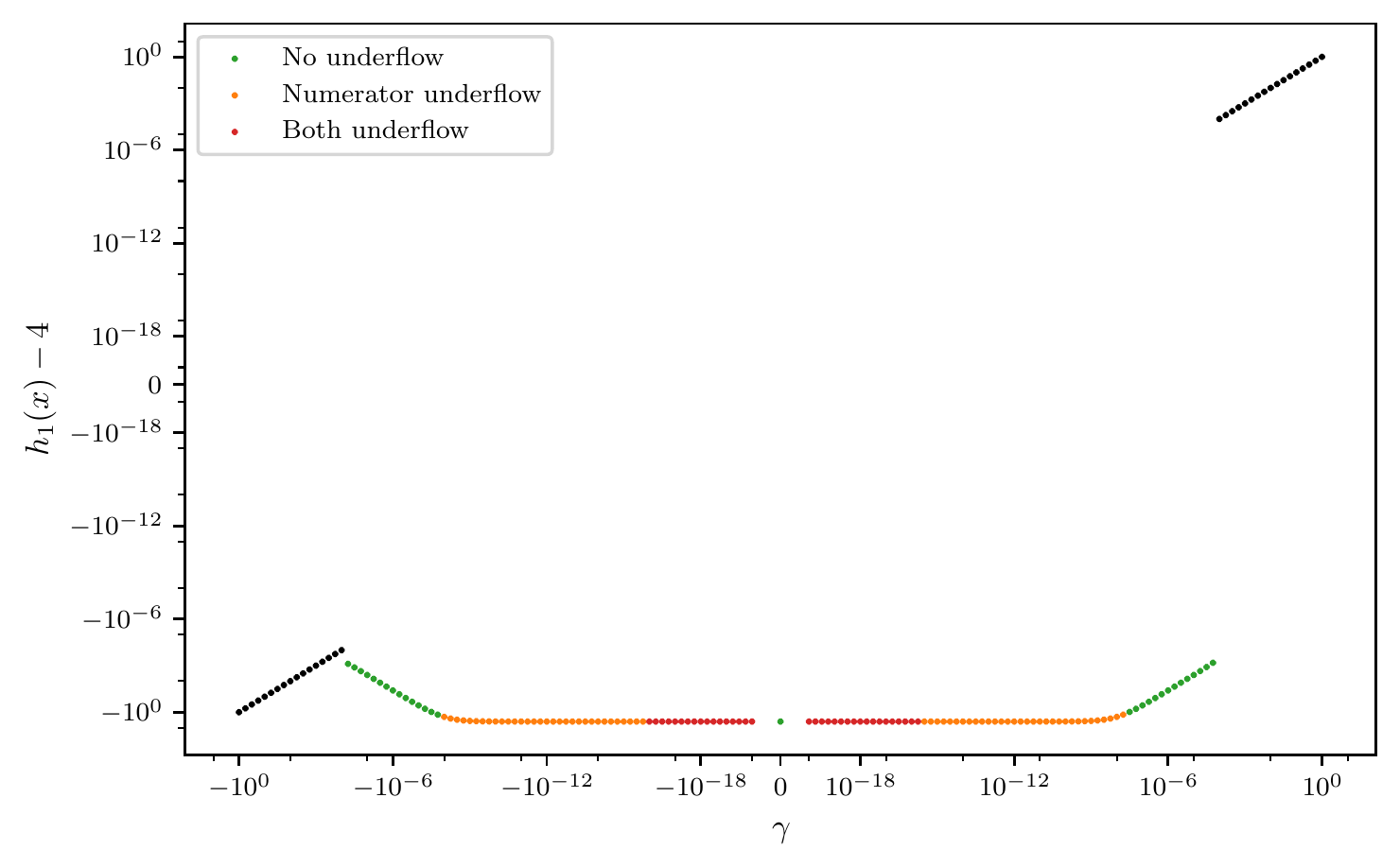}
	\caption{
 Same data as Figure \ref{fig:h_1_84_y}, except that the leading digit of 4 is removed and the dependent axis uses a log scale. This elucidates the linearity in the black region and the discontinuity between the black and green regions.}
	\label{fig:h_1_84_y-4}
\end{figure}

A common strategy for removing the \nans\ (and overflow) is to bound the denominator away from zero.  A typical way to accomplish this is to define
\begin{equation}
h_1(x; \epsilon, \delta)
=
\begin{cases}
\frac{f(x)}{g(x)+\signgx\epsilon}&\!\text{if } 2 - \delta <  x < 2+\delta\\
\frac{f(x)}{g(x)}&\text{otherwise} \\
\end{cases}
\label{eq:h1}
\end{equation}
where $\sign$ is the sign function and $\sign(0)$ is chosen arbitrarily.  Here, $\epsilon > 0$ and $\delta > 0$ are constants.  Although $h_1(x;\epsilon,\delta)$ is necessarily discontinuous, the discontinuity can be somewhat hidden by choosing $\delta$ large enough and $\epsilon$ small enough such that $\epsilon \ll |g(2 \pm \delta)|$. 
See Figures \ref{fig:h_1_84_y} and \ref{fig:h_1_84_y-4}.
The $O(1)$ errors shown in Figures \ref{fig:h_1_84_y} and \ref{fig:h_1_84_y-4} are well known in numerical analysis communities (see e.g.,~\cite{heath2018scientific, golub2013matrix, burden2015numerical, cheney2012numerical, atkinson1991introduction}) and typically addressed by refactoring code so that $h$ does not utilize two separate functions; instead, $h$ would be implemented as $\hat{h}(x)=x+2$.

\begin{figure}[h]
	\centering
\includegraphics[width=0.9\linewidth]{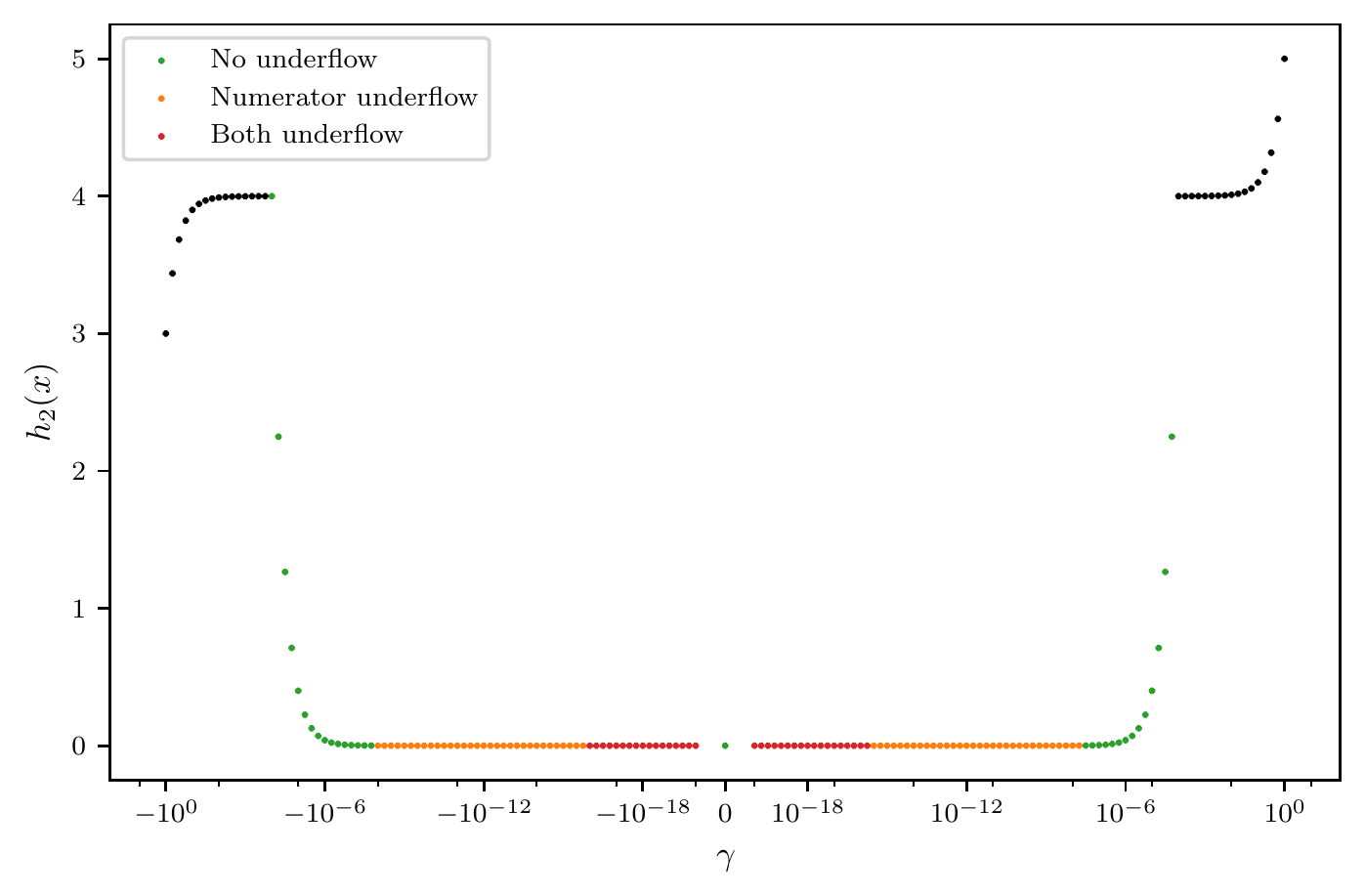}
	\caption{
Plot of $h_2(2+\gamma)$ with $\epsilon = 10^{-4}$; in other words, a plot of $\hat{h}_1(2+\gamma)$ with $\epsilon = 10^{-4}$ chosen to match $\delta = 10^{-4}$ for continuity.  Unlike in Figure~\ref{fig:h_1_84_y} where $g(x)$ slowly vanishes providing for a gradual transition between the black and green regions, the transition is more abrupt here even though the function is continuous; however, a more fair comparison would use $\epsilon = 10^{-8}$ in this figure, although that would place the transition in the yellow region (which we would like to avoid for the sake of exposition).
 }
	\label{fig:h3}
\end{figure}

If $\epsilon$ is chosen large enough and $\delta$ is chosen small enough, then $g(x)$ can be ignored to obtain
\begin{equation}
\hat{h}_1(x; \epsilon, \delta)
=
\begin{cases}
\frac{f(x)}{\signgx\epsilon} & \text{if } 2 - \delta < x < 2+\delta\\
\frac{f(x)}{g(x)} & \text{otherwise} \\
\end{cases} 
\label{eq:h2} 
\end{equation}
for simplicity.
In this case, the discontinuity can be removed by choosing $\delta$ so that $g(2 \pm \delta) = \sign(g(2 \pm \delta)) \epsilon$.  
This continuous case can be written with $g(x)$ in the conditional as 
\begin{equation}
h_2(x; \epsilon)
=
\begin{cases}
\frac{f(x)}{\signgx\epsilon} & \text{if } |g(x)|<\epsilon\\
\frac{f(x)}{g(x)} & \text{otherwise} \\
\end{cases} 
\label{eq:h3} 
\end{equation}
removing $\delta$; moreover, one can write 
\begin{equation}
\hat{h}_2(x; \epsilon)
=
\begin{cases}
\frac{f(x)}{-\epsilon} & \text{if } -\epsilon < g(x) < 0 \\
\frac{f(x)}{\epsilon} & \text{if } 0 < g(x) < \epsilon\\
\frac{f(x)}{g(x)} & \text{otherwise} \\
\end{cases} \label{eq:h4}
\end{equation}
in order to remove the sign function (although an arbitrary choice is still required when $g(x) = 0$).  
See Figure~\ref{fig:h3}.

\section{Computing Derivatives}
\begin{figure}[!h]
	\centering
    \includegraphics[width=0.9\linewidth]{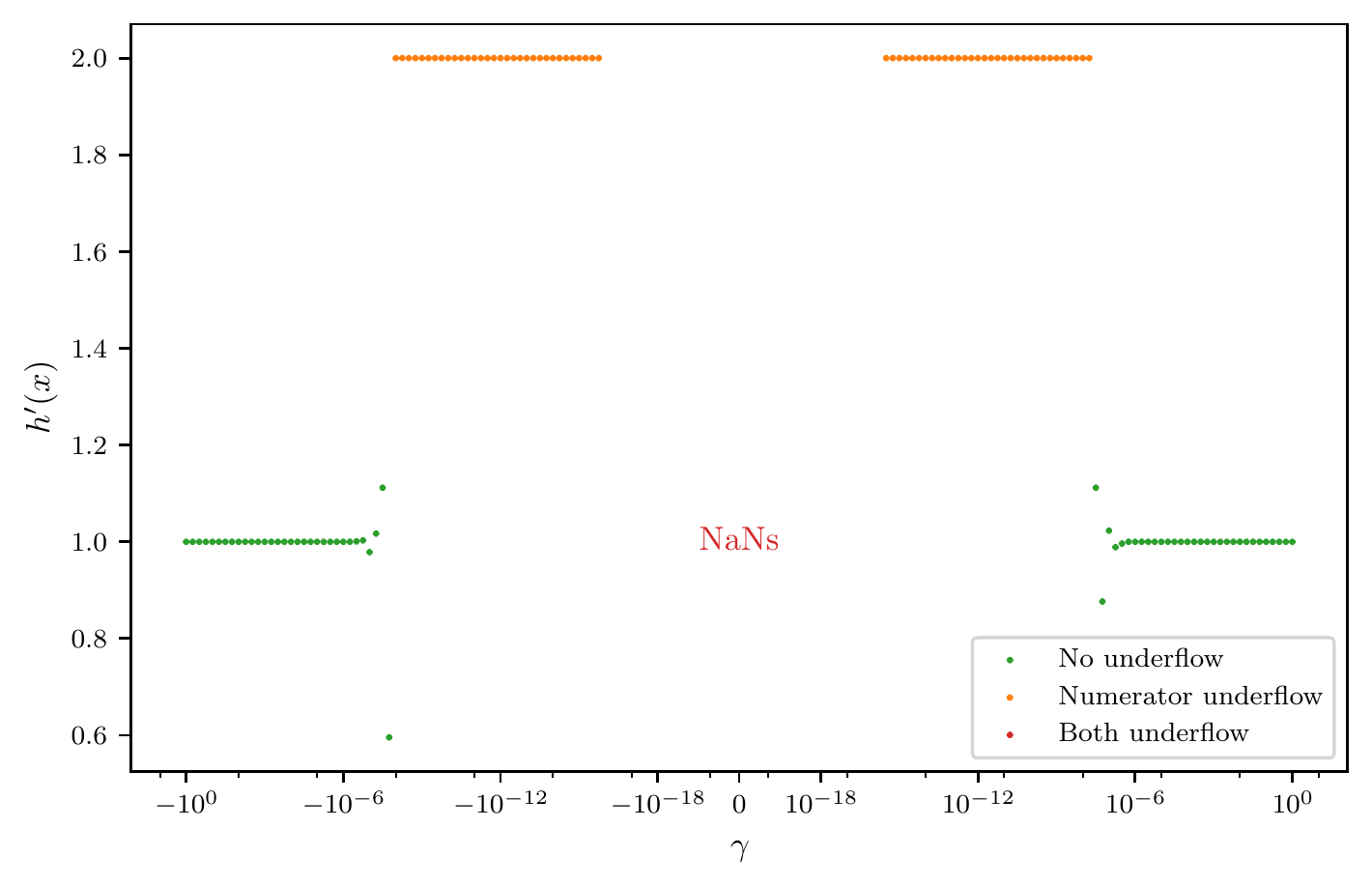}
	\caption{Although $h'(x) = 1$ identically, a value of 2 is obtained when $4 + 4\gamma + \gamma^2$ underflows to be $4 + 4\gamma$ because $\gamma^2 \ll 4$.  In addition, the errors near the boundary of the green region are caused by $4 + 4\gamma + \gamma^2$ partially underflowing to $4 + 4\gamma + \alpha \gamma^2$; as can be seen in equation~\ref{eq:hprime_mid}, this leads to a derivative that looks like $2-\alpha$.  Recall (from Figure~\ref{fig:h-4}) that one would expect $0 < \alpha < 1$ for rounding down and $1 < \alpha < 2$ for rounding up.}
	\label{fig:hp}
\end{figure}

Analytically, $f'(x) = 2x$, $g'(x) = 1$, and $h'(x)=1$ everywhere (after removing the singularity to reduce $h(x)$ to $x+2$). 
Backpropagation will utilize
\begin{subequations}
\begin{align}
dh &= \left(\frac{\partial h}{\partial f} \frac{\partial f}{\partial x} + \frac{\partial h}{\partial g} \frac{\partial g}{\partial x}\right)dx \\
&= \left( \frac{1}{g(x)} \frac{\partial f}{\partial x} -\frac{f(x)}{g(x)^2} \frac{\partial g}{\partial x}\right)dx \label{eq:dh_mid}\\
&= \left( \frac{2x}{x-2}-\frac{x^2-4}{(x-2)^2} \right) dx
\label{eq:dh}
\end{align} 
\end{subequations}
to compute derivatives.  When $2+\gamma$ underflows to be 2, $h'(2+\gamma) \approx \frac{4}{0} -\frac{0}{0}$ which gives \nans; otherwise, 
\begin{equation} \label{eq:hprime_mid}
h'(2+\gamma) = \frac{4+2\gamma}{\gamma} - \frac{(4 + 4\gamma + \gamma^2 ) - 4}{\gamma^2}
\end{equation}
can lead to 
\begin{equation}\label{eq:hprime_fin}
h'(2+\gamma) \approx \frac{4+2\gamma}{\gamma} - \frac{(4 + 4\gamma) - 4}{\gamma^2} = 2 \neq 1
\end{equation}
when $4 + 4\gamma + \gamma^2$ underflows to be $4 + 4\gamma$ because $\gamma^2 \ll 4$. 
Alternatively, the quotient rule would combine equation~\ref{eq:hprime_mid} into a single fraction, which has no bearing on the underflow and thus also leads to the final result in equation~\ref{eq:hprime_fin}.  See Figure \ref{fig:hp} for the results obtained using PyTorch (identical results were obtained using Tensorflow).

To analyze the derivative of $h_1(x)$, one can rewrite equation~\ref{eq:dh_mid} as, 
\begin{equation}
\label{eq:dh1}
dh_1 = \left( \frac{2x}{g_1(x)}  -\frac{x^2-4}{g_1(x)^2}g_1'(x)\right)dx
\end{equation}
with
\begin{equation}
\label{eq:g1}
g_1(x; \epsilon, \delta)
=
\begin{cases}
g(x)+\signgx\epsilon&\! \text{if } 2 - \delta <  x < 2+\delta\\
g(x)&\text{otherwise} \\
\end{cases}
\end{equation}
since the numerator remains unchanged.  Since $\sign'$ is identically zero almost everywhere, $g_1'(x) = g'(x) = 1$.  Note that attempts to smooth $\sign$ lead to $\sign$ taking on values of zero when $g$ is zero, defeating the purpose of adding $\signgx \epsilon$ to the denominator.

\begin{figure}[h]
	\centering
    \includegraphics[width=0.9\linewidth]{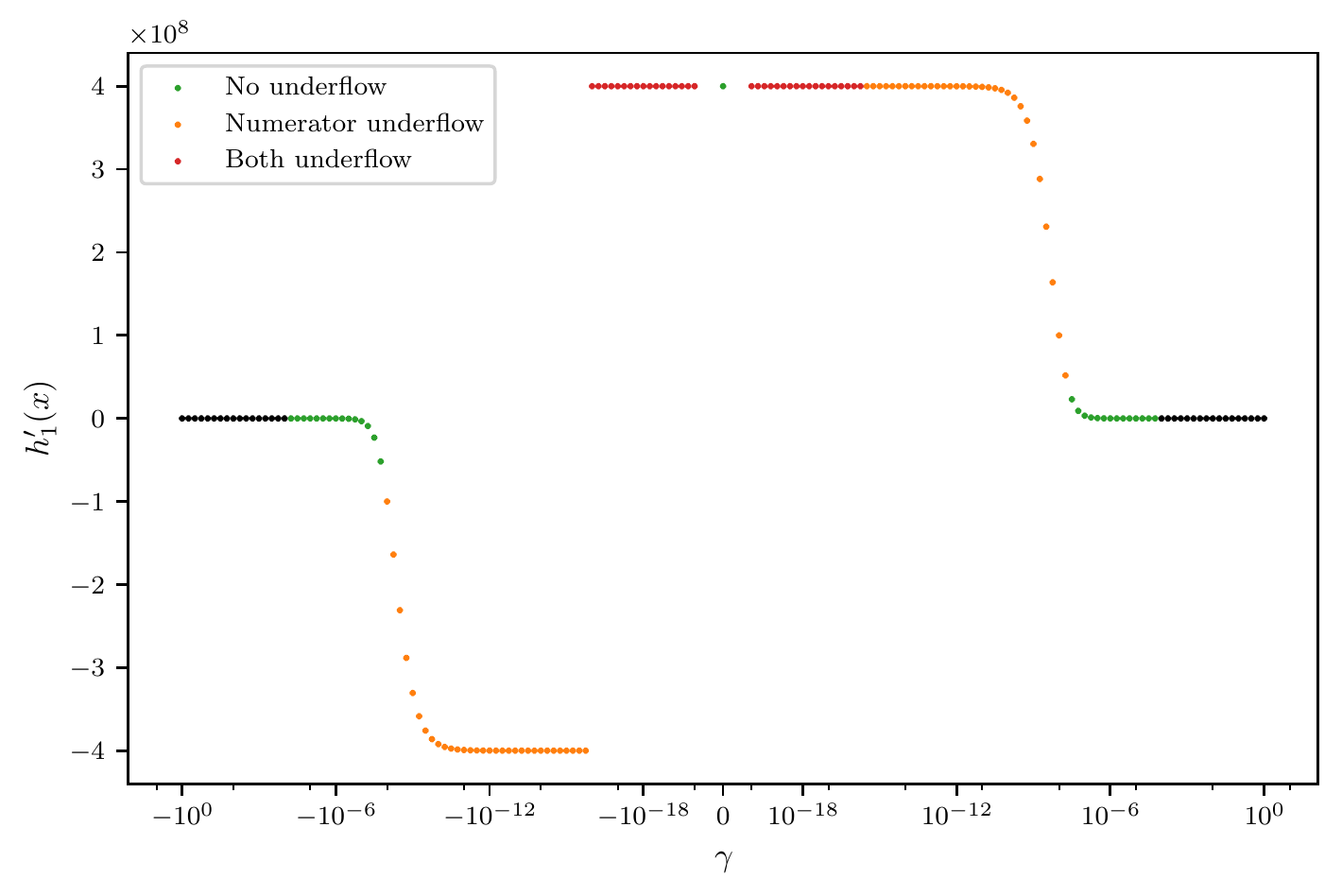}
	\caption{
As discussed before and after equation \ref{eq:h1p}, $h_1'$ has spurious derivative values on the order of $\frac{4}{\epsilon}$ which is $4\times 10^8$ for $\epsilon=10^{-8}$.
Note that $\sign(g(2+\gamma)) = \sign(0)$ in the red region where $2+\gamma$ underflows to $2$.
This means that all of the red points will agree in sign based on the arbitrary choice of $\sign(0)$;
obviously, we chose $\sign(0)=1$.
 }
	\label{fig:h1p}
\end{figure}

Outside of $x \in (2-\delta,2+\delta)$, the behavior of $h_1$ matches the behavior of $h$;  thus, we assume that $\delta$ is chosen large enough to encompass all of the degenerate behavior with the perturbed denominator.  When $2+\gamma$ underflows to be 2, $h_1'(2+\gamma) \approx  \frac{4}{\sign (0)\epsilon} - \frac{0}{\epsilon^2}$; otherwise, 
\begin{subequations}
\label{eq:h1p}
\begin{align}
h_1'(2+\gamma)
\label{eq:h1pa}
&\approx \frac{4+2\gamma}{\gamma+\sign (\gamma) \epsilon} - \frac{(4 + 4\gamma)-4}{(\gamma+\sign (\gamma) \epsilon)^2}\\ 
&\approx \frac{2\gamma^2 + (4 +2\gamma )\sign(\gamma) \epsilon}{(\gamma+\sign (\gamma) \epsilon)^2}
\end{align}
\end{subequations}
when $4+4\gamma + \gamma^2$ underflows to be $4 + 4\gamma$.  Note that $h_1'(2+\gamma) \to 2$ as $\epsilon \to 0$, consistent with equation~\ref{eq:hprime_fin} as expected; however, $h_1'(2+\gamma) \to \frac{4}{\sign (\gamma)\epsilon}$ as $\gamma \to 0$.  See Figure~\ref{fig:h1p} for the results obtained using PyTorch (identical results were obtained using Tensorflow).

\begin{remark}
Although the typical approach of adding a small number to the denominator in order to avoid \nans\ and overflow leads to $O(1)$ errors (e.g. $h_1(2) = 0$ when $h(2) = 4$), backpropagation through the perturbed formulas leads to unbounded errors (e.g. $h_1'(2) = \frac{4}{\sign (0)\epsilon}$ when $h'(2) = 1$).  Unfortunately, practitioners typically choose $\epsilon$ to be as small as possible.
\end{remark}

\begin{figure}[!h]
	\centering
    \includegraphics[width=0.9\linewidth]{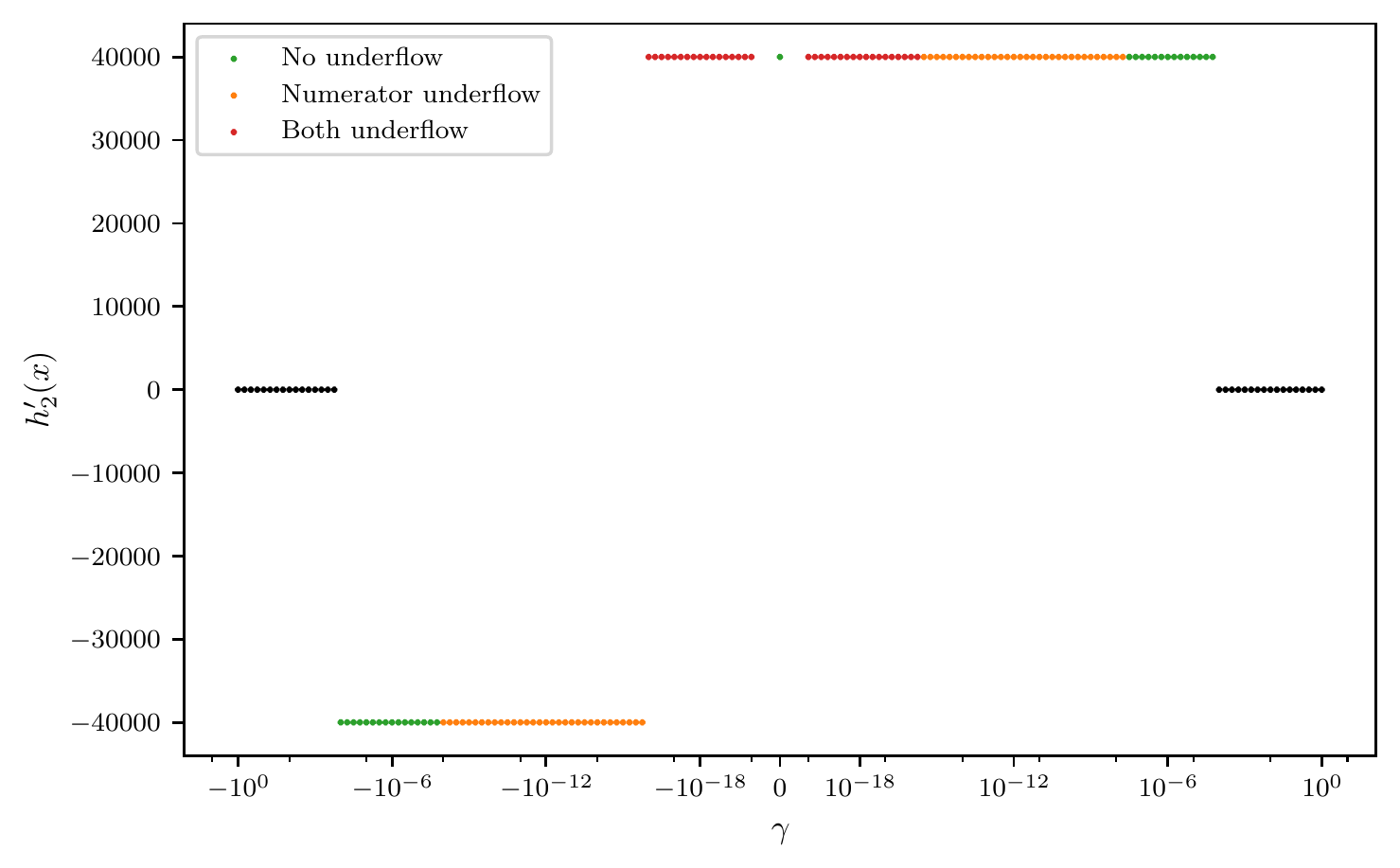}
	\caption{
    As discussed after equation \ref{eq:h2p}, $h_2'$ has spurious derivative values on the order of $\frac{4}{\epsilon}$ which is $4\times 10^4$ for $\epsilon=10^{-4}$;
    moreover, the derivatives blow up immediately (instead of gradually, as in Figure 7).
    }
	\label{fig:h3p}
\end{figure}

To analyze the derivative of $h_2(x)$, one can replace $g_1$ with
\begin{equation}
g_2(x; \epsilon)
=
\begin{cases}
\signgx\epsilon&\! \text{if } |g(x)| < \epsilon \\
g(x)&\text{otherwise} \\
\end{cases}
\label{eq:g2}
\end{equation}
in equation~\ref{eq:dh1}.  
In the region of interest where $|g(x)| < \epsilon$, $g_2'(x) = 0$.  
When $2+\gamma$ underflows to be 2, $h_2'(2+\gamma) \approx \frac{4}{\sign(0)\epsilon} - \frac{0}{\epsilon^2} \cdot 0$; otherwise, 
\begin{equation}
\label{eq:h2p}
h_2'(2+\gamma)
= \frac{4+2\gamma}{\sign(\gamma)\epsilon}
\end{equation}
whenever $|g(x)| < \epsilon$.  Note that any underflow in $f$ is unimportant since the result is robustly divided by $\epsilon^2$ before being multiplied by 0. 
Once again, $h_2'(2+\gamma)\to\frac{4}{\sign(\gamma)\epsilon}$ as $\gamma\to0$;
however, it happens more quickly in $h_2'$ (than in $h_1'$) because the second term in equation \ref{eq:dh1} is no longer present to partially cancel the first term (as it does in equation \ref{eq:h1pa}). 
See Figure~\ref{fig:h3p} for the results obtained using PyTorch (identical results were obtained using Tensorflow). 

\section{Conclusion}
As we have shown, the basic software infrastructures make the false assumption that expressions can be evaluated in nested fashion using the chain rule when the correct order of operations requires a cancelling of expressions before evaluating (as is taught in basic calculus when students learn about limits).
Although practitioners typically implement hacks into their codes in order to avoid division by zero, this leads to well-known $O(1)$ errors in the perturbed functions; unfortunately, the derivatives of those functions can have errors that are unbounded. The community does not seem to be concerned with these difficulties; rather, they over-embrace the modularity of the various software frameworks believing in the correctness of the derivatives that they produce.

Although this paper addressed the problems occurring with the very simple case of division, the lack of the ability to simplify expressions in object oriented code also causes problems with multiplication (e.g.~when $0\cdot\infty$ should be finite), addition/subtraction (e.g.~when $\infty-\infty$ should be finite), and many other operations.

Assuming that one takes precautions to avoid \texttt{NaN}s, it is important to consider what an erroneously large derivative (that blows up) will do when solving inverse/control problems or training neural networks.  A large value of the derivative in one entry of a gradient causes all of the other entries to vanish (comparatively), making the gradient erroneously point in the direction of the large derivative.  This leads to poor search directions in the optimization for first order methods, and (perhaps even more devastatingly) a corrupted Hessian for second order methods.  In turn, this undoes all of the hard work put into designing loss functions, regularizers, etc., in hopes of finding good parameters for the neural network.


\clearpage

\clearpage

\clearpage

\clearpage
\bibliographystyle{unsrtnat}  
\bibliography{refs}


\end{document}